\documentclass[journal]{IEEEtran}
\usepackage{amsfonts}
\usepackage{cite,graphicx,amsmath,amsthm}
\usepackage{fancyhdr}
\usepackage{dsfont}
\usepackage{array,color}
\usepackage{bm}
\usepackage{float}
\usepackage{algpseudocode}
\usepackage{multirow}
\usepackage{booktabs}
\usepackage{makecell}
\usepackage{hyperref}
\allowdisplaybreaks[4]
\usepackage{amsmath,amssymb, graphicx, epstopdf,cite,enumerate,booktabs, setspace, float,stfloats,bm, multirow, lscape, caption, subcaption, graphbox, soul, color, xcolor, hyperref}

\usepackage{diagbox}
\hypersetup{hidelinks}
\usepackage{cite, amsmath,amssymb, bm}
\usepackage{bbding}
\usepackage{amsthm}
\usepackage[normalem]{ulem}
\usepackage[linesnumbered,ruled]{algorithm2e}

\newtheorem{lemma}{Lemma}

\newtheorem{proposition}{Proposition}

\newtheorem{corollary}{Corollary}

\newtheorem{property}{Property}

\newtheorem{remark}{Remark}

\newtheorem{claim}{Claim}

\begin{document}

\title{
\vspace{-0.2in}
\LARGE Robotic Sensor Network: Achieving Mutual Communication Control\\Assistance With Fast Cross-Layer Optimization
}

\author{
Zhiyou Ji$^{*}$, Yujie Wan$^{*}$, Guoliang Li, Shuai Wang$^{\dag}$, Kejiang Ye,\\Derrick Wing Kwan Ng,~\emph{Fellow, IEEE}, and 
Chengzhong Xu,~\emph{Fellow, IEEE}
\vspace{-0.31in}
\thanks{
This work was supported by the National Natural Science Foundation of China (Grant No. 62371444), the Shenzhen Science and Technology Program (Grant No. RCYX20231211090206005), the Science and Technology Development Fund of Macao S.A.R (FDCT) (No. 0081/2022/A2), and the Guangdong Basic and Applied Basic Research Project (No. 2021B1515120067).

Zhiyou Ji, Shuai Wang, and Kejiang Ye are with the Shenzhen Institute of Advanced Technology, Chinese Academy of Sciences, Shenzhen, China.
Zhiyou Ji is also with the University of Chinese Academy of Sciences, China. 
Yujie Wan is with the Southern University of Science and Technology, Shenzhen, China, and is also with the Shenzhen Institute of Advanced Technology, Chinese Academy of Sciences, Shenzhen, China.
Guoliang Li and Chengzhong Xu are with the State Key Laboratory of IOTSC, Department of Computer and Information Science, University of Macau, Macau, China.
Derrick Wing Kwan Ng is with the School of Electrical Engineering and Telecommunications, the University of New South Wales, Australia.

Corresponding author: Shuai Wang ({\tt\footnotesize s.wang@siat.ac.cn}). 

$^*$These authors contribute equally. 
}
}
\maketitle

\begin{abstract}
Robotic sensor network (RSN) is an emerging paradigm that harvests data from remote sensors adopting mobile robots. 
However, communication and control functionalities in RSNs are interdependent, for which existing approaches become inefficient, as they plan robot trajectories merely based on unidirectional impact between communication and control.
This paper proposes the concept of mutual communication control assistance (MCCA), which leverages a model predictive communication and control (MPC$^2$) design for intertwined optimization of motion-assisted communication and communication-assisted collision avoidance. 
The MPC$^2$ problem jointly optimizes the cross-layer variables of sensor powers and robot actions, and is solved by alternating optimization, strong duality, and cross-horizon minorization maximization in real time. 
This approach contrasts with conventional communication control co-design methods that calculate an offline non-executable trajectory.
Experiments in a high-fidelity RSN simulator demonstrate that the proposed MCCA outperforms various benchmarks in terms of communication efficiency and navigation time.
\end{abstract}
\vspace{-0.06in}

\vspace{-0.25in}
\section{Introduction}

Edge intelligence exploits multi-modal data generated at distributed sensors for model training and inference \cite{zhou2019edge}. However, it suffers from the conflict between large data volumes and limited communication ranges, exacerbated by low transmit power of sensors.
Robotic sensor network (RSN) \cite{ma2012tour,chen2021ugv} offers a promising solution to mitigate this issue, by collecting data from sensing devices adopting mobile robots.
In RSN, the position of the wireless transceiver is controllable, representing an additional communication system variable to be optimized \cite{licea2024when}. 
This is different from conventional sensor networks, where the position of the transceiver remains static \cite{wang2019backscatter}. 

To satisfy the evolving requirements of RSN, it is necessary to integrate communication and motion features for joint optimization. This ignites extensive studies on 
motion-assisted communication (MAC) \cite{zhou2020learning,wang2019backscatter,guo2021uav,ali2018motion,yan2023communication}. 
However, communication also assists motion planning, since it allows obstacle states to be transmitted from external sensors to the ego-robot to facilitate collision avoidance. 
Existing MAC approaches overlook such mutual dependency between communication and control\footnote{Existing works on communication-assisted planning, e.g., \cite{jasontits}, do not consider how the robot motion impacts communication performance.}, leading to inefficient RSN designs.

Furthermore, existing MAC methods \cite{zhou2020learning,wang2019backscatter,guo2021uav,ali2018motion,yan2023communication} mainly focus on calculating a communication-friendly trajectory in an offline manner, due to the time-consuming optimization procedure. However, adopting such pre-planned trajectories for robot control often results in discrepancies between the intended path and robot's actual environments, primarily due to the need to avoid collisions with dynamic obstacles. 
This calls for fast cross-layer optimization algorithms for RSNs, which has not been well explored in current literature.

To fill the gap, this paper proposes a mutual communication control assistance (MCCA) framework, which leverages a model predictive communication and control (MPC$^2$) design for intertwined optimization of motion-assisted communication and communication-assisted collision avoidance.
The MPC$^2$ problem jointly optimizes the cross-layer variables of sensor powers and robot actions, and is solved by alternating optimization, strong duality, and cross-horizon minorization maximization (MM) in real time. 
This is in contrast to conventional MAC methods \cite{zhou2020learning,wang2019backscatter,guo2021uav,ali2018motion,yan2023communication} that calculate an offline non-executable trajectory.
To evaluate the proposed MCCA, we develop a high-fidelity platform based on robot operation system (ROS) and car learning to act (CARLA), which is closer-to-reality compared to numerical simulators adopted in \cite{zhou2020learning,wang2019backscatter,guo2021uav,ali2018motion,yan2023communication}. 
It is found that the communication efficiency gain achieved by MCCA over existing MAC and shape-based collision avoidance (SCA) can exceed over $10\%$.
Moreover, the collision avoidance capability of MCCA is shown to outperform MAC in an indoor navigation challenge, which demonstrates the benefits brought by communication to control.
Lastly, by adjusting a hyper-parameter in the cost function, the proposed MCCA also achieves the desired fairness among different sensors, which demonstrates the benefits of adopting cross-layer optimization.

\emph{Notation}.
Italic letters, simple bold letters, capital bold letters, and curlicue letters represent scalars, vectors, matrices, and sets, respectively.
$\nabla f$ represents the gradient of a function $f$, and $\mathbb{E}(\cdot)$ represents the expectation of a random variable.

\vspace{-0.12in}
\section{System Model}\label{section2}

We consider an RSN system shown in Fig. 1, which consists of $1$ robot collector and $K$ sensor devices, operating in an environment with $M$ obstacles.
The task is to generate a collision-free executable path (marked in red) while collecting the most data from the sensors. 
To prevent the robot from moving off the route, a global coarse path (marked in blue) is provided, which is represented by a list of waypoints $\mathcal{W}=\{\mathbf{w}_1, \mathbf{w}_2,\cdots\}$.\footnote{The coarse path is a naive straight-line generated by graph \cite{wang2019backscatter}.}

\begin{figure}[t]
    \centering
    \includegraphics[width=0.45\textwidth]{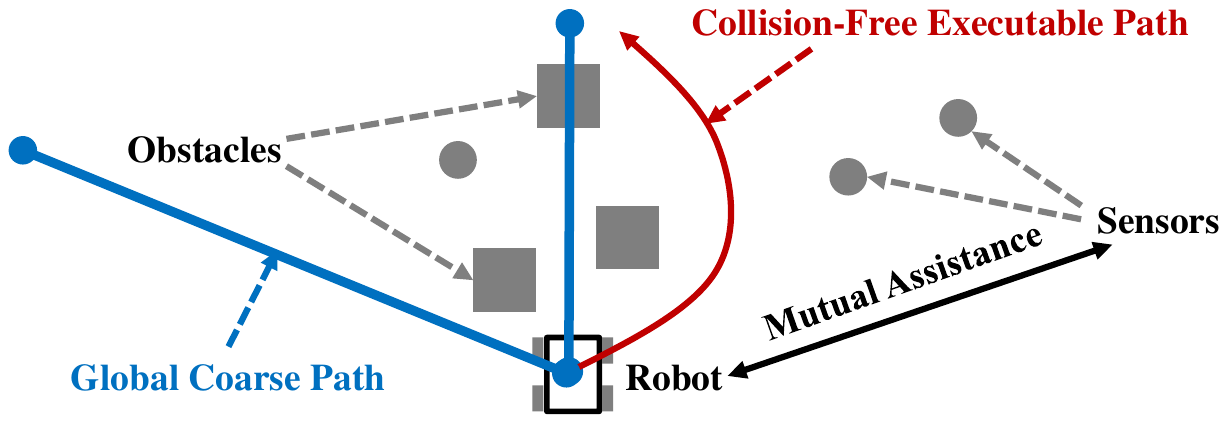}
    \caption{System model of RSN, where blue line denotes global coarse path, red line denotes collision-free executable path, grey boxes denote obstacles, grey balls denote sensors.}
    \vspace{-0.2in}
    \label{fig:model}
\end{figure}

\subsubsection{Robot Control}
To cope with dynamic environments, the robot control is required to operate in a receding-horizon fashion that replans at a high frequency (e.g., $10\,$Hz).
Each local horizon is divided into $H$ time slots with $\mathcal{H}=\{0,\cdots,H-1\}$, and 
the time step between consecutive slots is $\tau$. At the $t$-th time slot ($t\in\mathcal{H}$), the robot state is denoted as $\mathbf{s}_{t}=(x_{t},y_{t},\theta_{t})$,
where $(x_{t},y_{t})$ and $\theta_{t}$ are position and orientation of the robot, respectively. The robot action is denoted as 
$\mathbf{u}_{t}=(v_{t},\psi_{t})$,
where $v_{t}$ and $\psi_{t}$ are the linear and angular velocities, respectively.
The robot control is subject to the state evolution model $\mathbf{s}_{t+1}=E(\mathbf{s}_{t},\mathbf{u}_{t})$, where $E$ is given by \cite{zhang2020optimization}: 
\begin{align}
    E\left(\mathbf{s}_{t},\mathbf{u}_{t}\right) = \mathbf{A}_{t}{\mathbf{s}_{t}} + \mathbf{B}_{t}{\mathbf{u}_{t}} + \mathbf{c}_{t},~\forall t,
\end{align}
and coefficient matrices $(\mathbf{A}_{t}$, $\mathbf{B}_{t}$, $\mathbf{c}_{t})$ are determined by Ackermann kinetics and defined in \cite[Eqn. 8--10]{han2023rda}.
The action vector is bounded by $\mathbf{u}_{\min } \preceq \mathbf{u}_{t} \preceq \mathbf{u}_{\max },~\forall t$, and $\mathbf{a}_{\min }  \preceq  {\mathbf{u}_{t+1}}
    -{\mathbf{u}_{t}}   \preceq \mathbf{a}_{\max },~\forall t$, where ${\mathbf{u}_{\min }}$ and ${\mathbf{u}_{\max }}$ are the minimum and maximum values of the control vector, respectively, and ${\mathbf{a}_{\min }}$ and ${\mathbf{a}_{\max }}$ are the associated minimum and maximum accelerations, respectively.

\subsubsection{Robot Communication} When the robot moves to a position $(x_{t},y_{t})$ at the $t$-th time slot, it will collect data from sensors for a duration of $\tau$.
Within this $\tau$, the defined radio frequency (RF) source at the $k$-th sensor transmits a symbol sequence $z_{k,t}$ with $\mathbb{E}[|z_{k,t}|^2]=p_{k}$, where $p_{k}$ is the transmit power of the $k$-th sensor. 
The received signal-to-noise ratio (SNR) at the robot is
$\mathsf{SNR}_k=G_{k,t}(x_t,y_t)p_{k}\sigma^{-2}$,
where $G_{k,t}$ is the uplink channel gain
from the $k$-th sensor to the robot collector,
and $\sigma^2$ is the power of the complex-valued Gaussian noise.
Denoting the position of the $k$-th sensor device as $\mathbf{z}_{k}=(a_k,b_k)$, 
by adopting the distance dependent path loss model, the channel can be calculated as \cite{wang2019backscatter,guo2021uav}
\footnote{
Short-range communication typically belongs to line-of-sight transmission. If non-LOS transmission is considered, we need to adjust $\varrho_0$ accordingly. 
}
\begin{align}
G_{k,t}(x_t,y_t)=
\varrho_0
\|x_t-a_k, y_t-b_k\|_2^{-\alpha},
\end{align}
where $\varrho_0$ is pathloss at a reference distance of $1\,$meter and $\alpha\in[2,5]$ is path loss exponent. The spectral efficiency in bps/Hz of the $k$-th sensor is 
\begin{align}
R_{k,t}(\mathbf{s}_t,p_k)=\mathrm{log}_2\left(1+\frac{\varrho_0
\|x_t-a_k, y_t-b_k\|_2^{-\alpha}p_{k}}{\sigma^2}\right).
\end{align}

\section{Mutual Communication Control Assistance}\label{section3}

\subsection{Motion-Assisted Communication Utility}

In the considered system, our task is to maximize the amount of data harvested from sensors, by planning the robot trajectory, involving variables $\{\mathbf{s}_{t}\}_{t=0}^H$ and $\{\mathbf{u}_{t}\}_{t=0}^H$, and the sensor transmit powers, involving variables $\{p_k\}_{k=1}^K$.
This is realized by maximizing the following motion-assisted communication utility \cite{evangelista2019fairness}
\begin{align}
&C_0\left(\{\mathbf{s}_{t}\}_{t=0}^{H},\{p_k\}_{k=1}^K\right)=
(1-\beta)
\sum_{t=0}^H
\tau
\sum_{k=1}^KB_k
R_{k,t}(\mathbf{s}_t,p_k)
\nonumber\\
&\quad\quad \quad\quad \quad\quad \quad\quad\quad \
+\beta
\sum_{t=0}^H
\tau
\min_{1\leq k\leq K}B_k
R_{k,t}(\mathbf{s}_t,p_k)
,
\end{align}
where $B_k$ in Hz is the bandwidth allocated to sensor $k$, and $\beta$ is a fairness hyper-parameter ranging from $0$ to $1$.

It can be seen that if $\beta=0$, $C_0$ represents the total amount of data harvested from all sensors, which guarantees overall performance. If $\beta=1$, $C_0$ represents the worst communication performance among all users, which guarantees user fairness.
By adjusting $\beta$ between $0$ and $1$, we can strike an effective balance between overall performance and user fairness according to different system requirements.

\vspace{-0.1in}
\subsection{Communication-Assisted Collision Avoidance}

The robot is not allowed to collide with any obstacle $m\in\mathcal{E}_t\cup\{\mathcal{S}_{j,t}\}_{j=1}^K$, where set $\mathcal{E}_t$ denotes the list of obstacles detected by ego-robot, and 
set $\mathcal{S}_{j,t}$ denotes the list of obstacles informed by the external sensor $j\in\{1,\cdots,K\}$. 
Each $\mathcal{S}_{j,t}$ depends on the communication power $p_j$ as follows:
\begin{equation}
\mathcal{S}_{j,t}(p_j)=
\left\{\begin{array}{ll}
\mathcal{O}_{j,t}, &\mathrm{if}~j\in\mathcal{A},
\frac{B_j}{H}\sum_{t=0}^H
R_{j,t}(\mathbf{s}_t,p_j) \geq D_0 \\
\emptyset, &\mathrm{otherwise}
\end{array}\right.,
\end{equation} 
where $\mathcal{O}_{j,t}$ denotes the list of obstacles detected by the external sensor $j$, $\mathcal{A}$ is a predetermined set of sensors that can assist collision avoidance, $D_0$ is the minimum data-rate for real-time sensing data exchange.
The ego-robot and external sensors adopt lidar point cloud and the sparsely embedded convolutional detection (SECOND) network to detect obstacles for obtaining $\mathcal{E}_t$ and $\mathcal{O}_{j,t}$ \cite{flcav}. 
The communication-assisted collision avoidance constraint is thus given by
\begin{align} 
& 
{\bf{dist}}_{m,t}\left(\mathbf{s}_{t}\right)
\geq d_{\mathrm{safe}},~\forall m\in\mathcal{E}_t\cup\{\mathcal{S}_{j,t}\}_{j=1}^K,\ \forall t,
\label{A1}
\end{align}
where ${\bf{dist}}_{m,t}\left(\mathbf{s}_{t}\right)$ is the distance between the robot and the $m$-th obstacle at time $t$, and $d_{\mathrm{safe}}>0$ is safety distance.

We adopt shape-distance model for $\bf{{dist}}_{m,t}$ based on set representation. The occupancy region of the $m$-th obstacle ${\mathbb{O}_{m,t}}$ is represented as a set $\mathbb{O}_{m,t}= \{ \mathbf{z}\in {\mathbb{R}^{3}}| {\mathbf{H}_{m,t}}\mathbf{z}\preceq 
     {\mathbf{h}_{m,t}}\}
$, where $\mathbf{H}_{m,t}$ and $\mathbf{h}_{m,t}$ are the rotation matrix and translation vector of all contour lines for the $m$-th obstacle, respectively.
The robot occupancy region with respect to state $\mathbf{s}_t$ is 
$\mathbb{G}_{t}(\mathbf{s}_t)=\{\mathbf{z}\in {\mathbb{R}^{3}}| \mathbf{G}_t(\mathbf{s}_t)\mathbf{z}\preceq 
     {\mathbf{g}}_t(\mathbf{s}_t)\}$, where $\mathbf{G}_t(\mathbf{s}_{t})$ is the rotation matrix related to $\theta_{t}$ and $\mathbf{g}_t(\mathbf{s}_{t})$ is the translation vector related to $\left(x_{t},{y_{t}}\right)$. 
To determine $\bf{{dist}}_{m,t}(\mathbf{s}_t)$, it is equivalent to compute the distances between any two points within 
$\mathbb{G}_{t}$ and 
$\mathbb{O}_{m,t}$, and then take the minimum, i.e.,
\begin{equation}
{\mathbf{dist}}_{m,t}({\mathbf{s}_{t}}) =
\min \{\|\mathbf{x}-\mathbf{y}\|_2 | \mathbf{x}\in\mathbb{G}_{t}(\mathbf{s}_t), \mathbf{y}\in \mathbb{O}_{m,t} \}.  \label{cac}
\end{equation}

\vspace{-0.1in}
\subsection{Model Predictive Communication and Control}

Based on Sections III-A and III-B, the $\mathsf{MPC}^2$ is formulated as the following cross-layer optimization problem: 
\begin{subequations}
\vspace{-0.1in}
\begin{align}
    (\mathsf{MPC}^2)
    &\max \limits_{\{\mathbf{s}_{t}, \mathbf{u}_{t}, p_{k}\}}~~
    C_0\left(\{\mathbf{s}_{t},p_k\}\right)
    - \rho\sum^{H}_{t=0} 
    \left\|\mathbf{s}_{t}-\mathbf{w}_{t}^\diamond\right\|^2  \label{utility} \\
    \text { s.t. }~~&\mathbf{s}_{t+1} = \mathbf{A}_{t}{\mathbf{s}_{t}} + \mathbf{B}_{t}{\mathbf{u}_{t}} + \mathbf{c}_{t},~\forall t, \label{dynamics}\\
    &\mathbf{u}_{\min } \preceq \mathbf{u}_{t} \preceq \mathbf{u}_{\max },~\forall t, \label{bounds}\\
    &\mathbf{a}_{\min }  \preceq  {\mathbf{u}_{t+1}}
    -{\mathbf{u}_{t}}   \preceq \mathbf{a}_{\max },~\forall t, \label{bounds2} \\
    &{\bf{dist}}_{m,t}\left(\mathbf{s}_{t}\right)
\geq d_{\mathrm{safe}},~\forall m\in\mathcal{E}_t\cup\{\mathcal{S}_{j,t}\},\ \forall t, \label{collision}\\
    & \frac{1}{K}\sum_{k=1}^Kp_k\leq P,\ p_k\geq 0,\ \forall k, \label{A6}
\end{align}
\end{subequations}
where an additional cost $-\rho\sum^{H}_{t=0}\left\|\mathbf{s}_{t}-\mathbf{w}_{t}^\diamond\right\|^2$ is introduced to $C_0$, which prevents the robot from moving off the route, with $\rho>0$ being a tunable parameter, and equation \eqref{A6} is for constraining the power consumption at the sensors, with $P$ being the power budget.
Problem $\mathsf{MPC}^2$ is nontrivial to solve due to the nonlinear coupling between cross-layer variables $\{\mathbf{s}_{t}, \mathbf{u}_{t}\}$ and $ \{p_{k}\}$, the nonconvex function $C_0$ in motion-assisted communication utility \eqref{utility}, and the implicit function ${\bf{dist}}_{m,t}$ in communication-assisted motion constraint \eqref{collision}. 

\section{Fast Cross-Layer Optimization}\label{section3}

\subsection{Alternating Optimization}

To resolve the coupling between $\{\mathbf{s}_{t}, \mathbf{u}_{t}\}$ and $ \{p_{k}\}$ in $\mathsf{MPC}^2$, we adopt alternating optimization that iterates between solving $\{\mathbf{s}_{t}, \mathbf{u}_{t}\}$'s subproblem and $\{p_{k}\}$'s subproblem. 
Given an initial guess of $\{p_{k}\}$ (e.g., $\{p_{k}=P\}$), in each iteration, $\{\mathbf{s}_{t}, \mathbf{u}_{t}\}$ is first optimized given fixed $\{p_{k}=p_{k}'\}$, and then $\{p_{k}\}$ is optimized by fixing $\{\mathbf{s}_{t}=\mathbf{s}_{t}', \mathbf{u}_{t}=\mathbf{u}_{t}'\}$ with the new values of $\{\mathbf{s}_{t}',\mathbf{u}_{t}'\}$ obtained from the $\{\mathbf{s}_{t}, \mathbf{u}_{t}\}$'s problem. 
The $\{p_{k}\}$'s subproblem is
\begin{subequations}
\begin{align}
(\mathsf{P}_1)~\max \limits_{\{p_{k}\}}~~&
    C_0\left(\{\mathbf{s}_{t}',p_k\}\right)  \\
    \text { s.t. }~~&
    \frac{1}{H}\sum_{t=0}^H B_k
R_{k,t}(\mathbf{s}_t',p_k) \geq D_0,~\forall k\in \mathcal{A}, \\
    & \frac{1}{K}\sum_{k=1}^Kp_k\leq P,\ p_k\geq 0,\ \forall k,
\end{align}
\end{subequations}
which is a convex optimization problem that is solved by domain-specific optimization software (e.g., CVXPY) \cite{diamond2016cvxpy}. On the other hand, the $\{\mathbf{s}_{t}, \mathbf{u}_{t}\}$'s subproblem is 
\begin{subequations}
\vspace{-0.1in}
\begin{align}
    (\mathsf{P}_2)~&\max \limits_{\{\mathbf{s}_{t}, \mathbf{u}_{t}\}}~~
    C_0\left(\{\mathbf{s}_{t},p_k'\}\right)
    - \rho\sum^{H}_{t=0} 
    \left\|\mathbf{s}_{t}-\mathbf{w}_{t}^\diamond\right\|^2  \\
    \text { s.t. }~~&\eqref{dynamics}, \eqref{bounds}, \eqref{bounds2} \\
    &{\bf{dist}}_{m,t}\left(\mathbf{s}_{t}\right)
\geq d_{\mathrm{safe}},~\forall m\in\mathcal{E}_t\cup\{\mathcal{S}_{j,t}\},\ \forall t,  \label{collision2}
\end{align}
\end{subequations}
which will be tackled in the subsequent parts.

\subsection{Duality-Based Optimization}

To tackle the function ${\bf{dist}}$ in \eqref{collision2}, it is first observed that problem \eqref{cac} is a linearly constrained quadratic programming problem, for which strong duality always hold. 
As such, we leverage this property to transform \eqref{cac} into its dual 
\begin{subequations}
\begin{align}
(\mathsf{D})
&\mathop{\mathrm{max}}_{\{\bm{\gamma}_{m,t},\bm{\phi}_{m,t}\}}~
 \bm{\gamma}_{m,t}^T{\mathbf{H}_{m,t}} -{\bm{\gamma}_{m,t}^T}{\mathbf{h}_{m,t}}- {\bm{\phi}_{m,t}^T}{\mathbf{g}}_t(\mathbf{s}_t) \\
&~~~~\mathrm{s.t.}~~
    {\left\| {{\mathbf{H}_{m,t}^T}\bm{\gamma}_{m,t} } \right\|} \leq 1, ~
    \bm{\gamma}_{m,t} \succeq \mathbf{0}, ~\bm{\phi}_{m,t} { \succeq}\mathbf{0}, \\
    &~~~~~~~~~~{\bm{\phi}_{m,t}^T}{\mathbf{G}_t(\mathbf{s}_t)}
    +{\bm{\gamma}_{m,t} ^T}{\mathbf{H}_{m,t}}=\mathbf{0},
\label{dual}  
\end{align}
\end{subequations}
where $\{\bm{\gamma}_{m,t}$, $\bm{\phi}_{m,t}\}$ are dual variables. 
By solving $\mathsf{D}$ we obtain the optimal solution $\{\widehat{\bm{\gamma}}_{m,t}$, $\widehat{\bm{\phi}}_{m,t}\}$, and \eqref{collision} becomes 
\begin{subequations}
\begin{align}
&\widehat{\bm{\gamma}}_{m,t}^T{\mathbf{H}_{m,t}} -{\widehat{\bm{\gamma}}_{m,t}^T}{\mathbf{h}_{m,t}}- {\widehat{\bm{\phi}}_{m,t}^T}{\mathbf{g}_t(\mathbf{s}_t)}\geq d_{\mathrm{safe}},~\forall t, \label{dual1}  
\\
&
{\widehat{\bm{\phi}}_{m,t}^T}{\mathbf{G}_t(\mathbf{s}_t)}
    +{\widehat{\bm{\gamma}}_{m,t} ^T}{\mathbf{H}_{m,t}}=\mathbf{0},~\forall t, 
 \label{dual2}  
\end{align}
\end{subequations}
where \eqref{dual1} is the dual-based collision avoidance constraint and \eqref{dual2} is the primal-dual equality constraint, which are both convex in ${\mathbf{s}_{t}}$.

\subsection{Cross-Horizon MM}

To tackle the nonconvexity of $C_0$ in $\mathbf{s}_t$, we propose to use the framework of MM, which constructs a sequence of concave lower bounds $\{{\Phi}_{k,t}\}$ on $\{R_{k,t}\}$ and replaces $\{R_{k,t}\}$ with $\{\Phi_{k,t}\}$ to obtain a sequence of surrogate problems.
More specifically, given any feasible solution 
$\{\mathbf{s}_{t}^\star, 
\mathbf{u}_{t}^\star\}$, to $\mathsf{P}_2$, define its surrogate function as
\begin{align}
&\Phi_{k,t}(\mathbf{s}_{t}|\mathbf{s}_{t}^\star)
=\mathrm{log}_2\Bigg(1+\frac{\varrho_0 p_{k}}{\sigma^2}\times
\Big[
2\|\mathbf{s}_{t}^\star
-[a_{k},b_{k}, \theta_t^\star]^T\|_2^{-\alpha}
\nonumber\\
&
-\|\mathbf{s}_{t}^\star
-[a_{k},b_{k}, \theta_t^\star]^T\|_2^{-2\alpha}
\left\|
x_{t}-a_{k}, y_{t}-b_{k}\right\|_2^{\alpha}\Big]
\Bigg),
\end{align}
and the following proposition can be established.

\begin{proposition}
$\{\Phi_{k,t}\}$ satisfy the following conditions:

\noindent(i) Concavity: $\Phi_{k,t}(\mathbf{s}_{t}|\mathbf{s}_{t}^\star)$ is concave in $\mathbf{s}_{t}$.

\noindent(ii) Lower bound: $\Phi_{k,t}(\mathbf{s}_{t}|\mathbf{s}_{t}^\star)\leq R_{k,t}(\mathbf{s}_{t})$.

\noindent(iii)  
$\Phi_{k,t}(\mathbf{s}_{t}^\star|\mathbf{s}_{t}^\star)=R_{k,t}(\mathbf{s}_{t}^\star)$ and $\nabla\Phi_{k,t}(\mathbf{s}_{t}^\star|\mathbf{s}_{t}^\star)=\nabla R_{k,t}(\mathbf{s}_{t}^\star)$.

\end{proposition}
\begin{proof}
Part (i) is proved by checking the semi-definiteness of Hessian of $\Phi_{k,t}$;
Part (ii) is proved based on 
$\frac{1}{x}\geq \frac{1}{y}-\frac{1}{y^2}(x-y)$ for any $(x,y)$;
Part (iii) is proved by computing the function and gradient values of  $\Phi_{k,t}$ and  $R_{k,t}$.
\end{proof}
Now, assuming that the solution at the $n$-th iteration is given by 
$\{\mathbf{s}_{t}^{[n]}, 
\mathbf{u}_{t}^{[n]}\}$, the following problem is considered at the $(n+1)$-th iteration:
  \begin{align}\label{Pn+1}
    &\max \limits_{\{\mathbf{s}_{t}, \mathbf{u}_{t}\}}~~
    (1-\beta)
    \sum_{h=0}^H\tau
\sum_{k=1}^KB_k\Phi_{k,t}(\mathbf{s}_{t}|
\mathbf{s}_{t}^{[n]}
) 
\nonumber\\
&
+\beta
\sum_{t=0}^H
\tau
\min_{1\leq k\leq K}B_k
\Phi_{k,t}(\mathbf{s}_{t}|
\mathbf{s}_{t}^{[n]}
)
-\rho\sum^{H}_{t=0} 
    \left\|\mathbf{s}_{t}-\mathbf{w}_{t}^\diamond\right\|^2
\nonumber \\
    &\text { s.t. }~~\eqref{dynamics},\eqref{bounds},\eqref{bounds2},\eqref{dual1},\eqref{dual2}.
    \end{align}
Based on part (i) of \textbf{Proposition 1}, the above problem is convex and can be readily solved.
Denoting its optimal solution as
$\{\mathbf{s}_{t}^*, 
\mathbf{u}_{t}^*\}$, we set
$\{\mathbf{s}_{t}^{[n+1]}=\mathbf{s}_{t}^*, 
\mathbf{u}_{t}^{[n+1]}=\mathbf{u}_{t}^*
\}$, and the process repeats with solving the $(n+2)$-th problem.
According to parts (ii) and (iii) of \textbf{Proposition 1} and \cite[Theorem 1]{sun2016majorization}, every limit point of sequence
$(
\{\mathbf{s}_{t}^{[0]},\mathbf{u}_{t}^{[0]}\},
\{\mathbf{s}_{t}^{[1]},\mathbf{u}_{t}^{[1]}\},
\cdots)$ is a Karush-Kuhn-Tucker solution to $\mathsf{P}_2$, as long as the starting point $\{\mathbf{s}_{t}^{[0]},\mathbf{u}_{t}^{[0]}\}$ is feasible to $\mathsf{P}_2$.

The worst-case complexity for solving (14) is
$\mathcal{O}(6H)^{3.5})$, as it involves $3H$ state variables and $3H$ action variables.
Therefore, the total complexity for solving $\mathsf{P}_2$ is $\mathcal{O}(\mathcal{I}\,(6H)^{3.5})$, where $\mathcal{I}$ is the number of iterations needed for MM to converge. 
To satisfy the requirement of real-time robot control, $\mathcal{I}$ can be reduced to $1$ by leveraging the cross-horizon MM. 
This is equivalent to \emph{conducting inexact one-shot MM in each horizon by spreading the MM iterations over consecutive horizons}.
Such cross-horizon MM leads to negligible performance loss in robot control, since the the state difference between consecutive horizons is small.

\section{Simulation Results}\label{section4}

This section evaluates the performance of the proposed MCCA. 
The sensor power budget is $P=1\,$mW and the communication bandwidth $B_k=0.1\,$MHz.
The noise power is $\sigma^2=-60\,$dBm (corresponding to $-110\,$dBm/Hz).
We adopt $\varrho_0=-30\,$dB, $\alpha=2$, $\tau=0.1\,$s, $d_{\mathrm{safe}}=1\,$m.
All simulations are implemented on a Ubuntu workstation with a $3.7$\,GHz AMD Ryzen 9 5900X CPU and an NVIDIA $4090$ GPU. 

We compare MCCA to the following baselines: 1) SCA \cite{han2023rda}, which ignores communication requirements;
2) MAC with point-based collision avoidance (PCA), which combines the design principle in
\cite{guo2021uav,ali2018motion} with PCA in \cite{jasontits}, where 
each obstacle, as well as ego-robot, is modeled as a sphere;
3) MAC with SCA, which combines the design principle in
\cite{guo2021uav,ali2018motion} with SCA in \cite{han2023rda}.

\begin{figure}[!t]
  \centering
  \begin{subfigure}[t]{0.215\textwidth}
      \centering
      \includegraphics[width=1\textwidth]{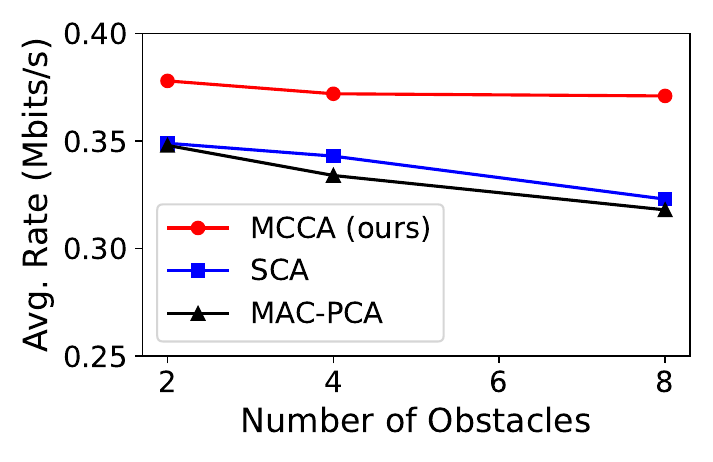}
      \caption{Communication efficiency.}
  \end{subfigure}
  \hfill
  \begin{subfigure}[t]{0.265\textwidth}
    \centering
    \includegraphics[width=1\textwidth]{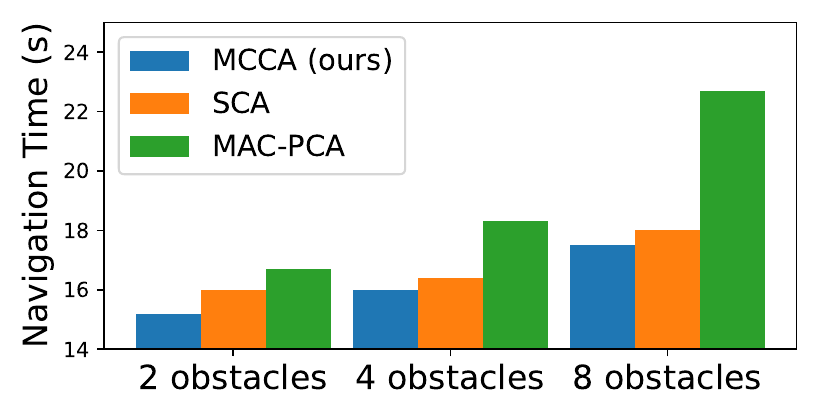}
    \caption{Navigation time.}
  \end{subfigure}
  \vspace{-0.05in}
  \caption{Quantitative comparison of different schemes.}
  \label{perf_los}
  \vspace{-0.15in}
\end{figure}

\begin{figure}[!t]
  \centering
  \begin{subfigure}[t]{0.15\textwidth}
    \includegraphics[width=1\textwidth]{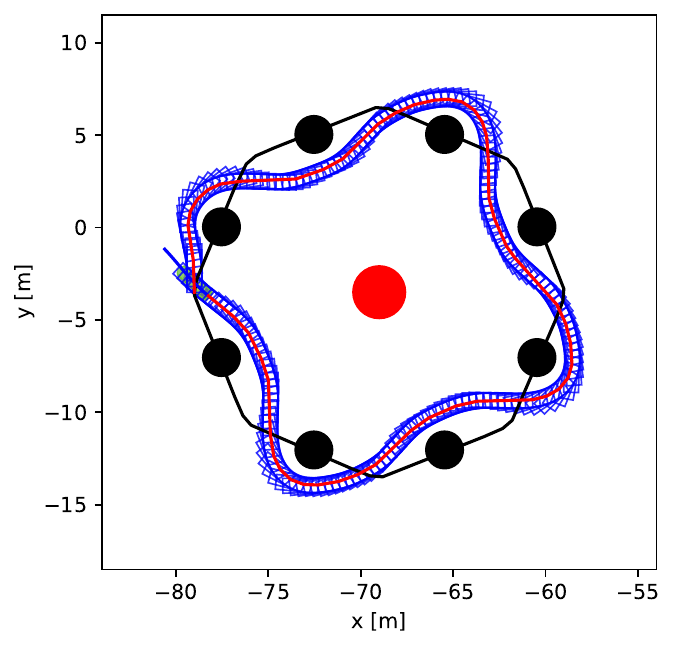}
      \vspace{-0.25in}
    \caption{MCCA (ours).}
  \end{subfigure}
      \hfill
  \begin{subfigure}[t]{0.15\textwidth}
      \includegraphics[width=1\textwidth]{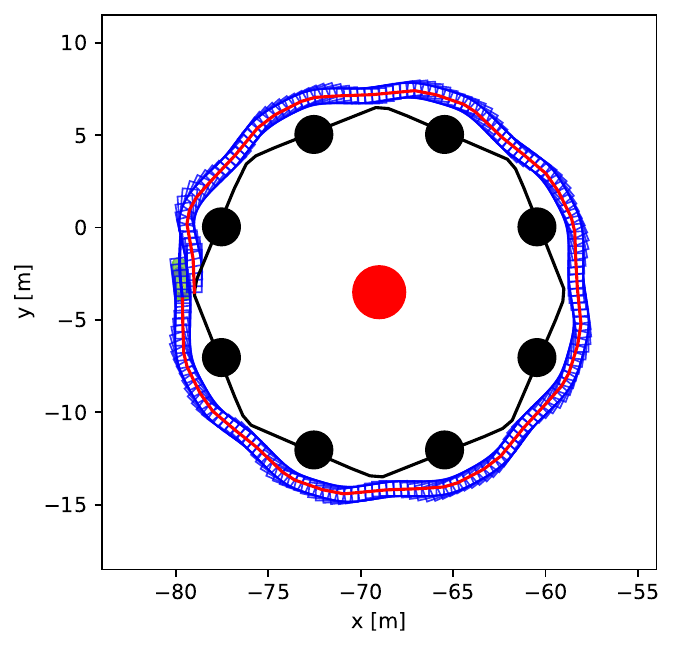}
      \vspace{-0.25in}
      \caption{SCA.}
  \end{subfigure}
  \hfill
  \begin{subfigure}[t]{0.15\textwidth}
    \includegraphics[width=1\textwidth]{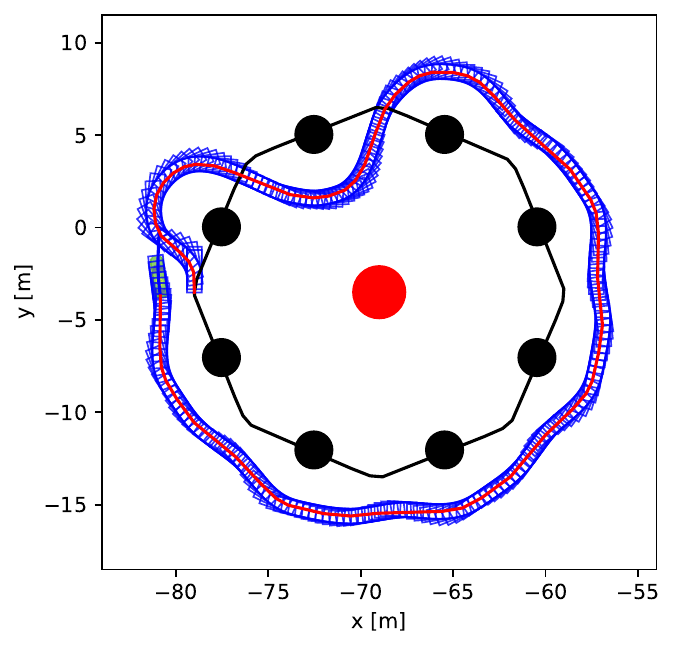}
      \vspace{-0.25in}
    \caption{MAC with PCA.}
  \end{subfigure}
    \begin{subfigure}[t]{0.15\textwidth}
    \includegraphics[width=1\textwidth]{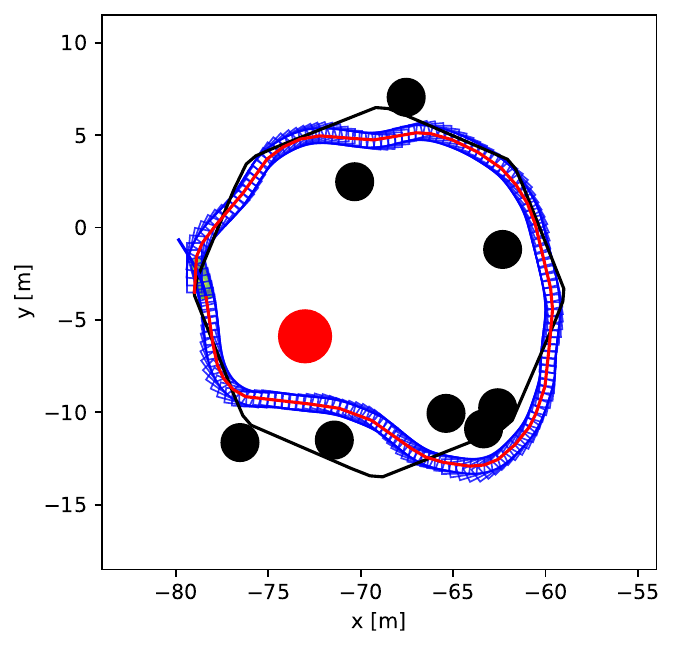}
          \vspace{-0.25in}
    \caption{MCCA (ours).}
  \end{subfigure}
      \hfill
  \begin{subfigure}[t]{0.15\textwidth}
      \includegraphics[width=1\textwidth]{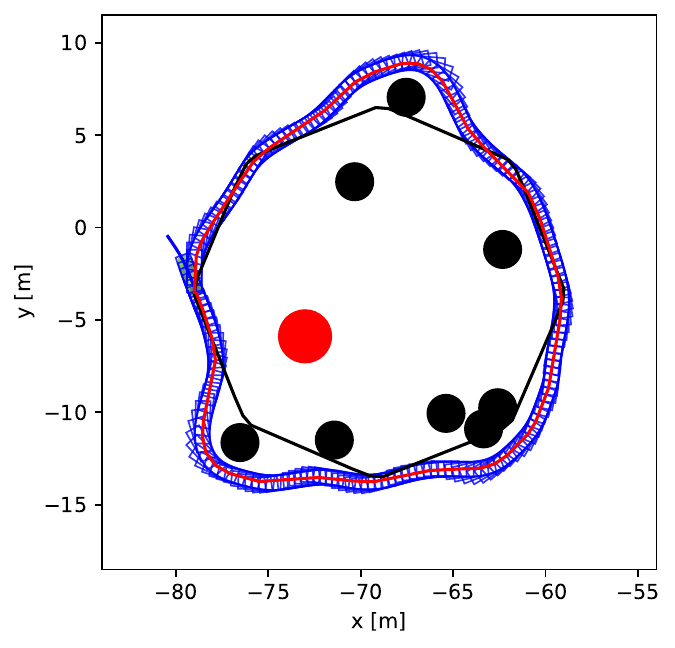}
            \vspace{-0.25in}
      \caption{SCA.}
  \end{subfigure}
  \hfill
  \begin{subfigure}[t]{0.15\textwidth}
    \includegraphics[width=1\textwidth]{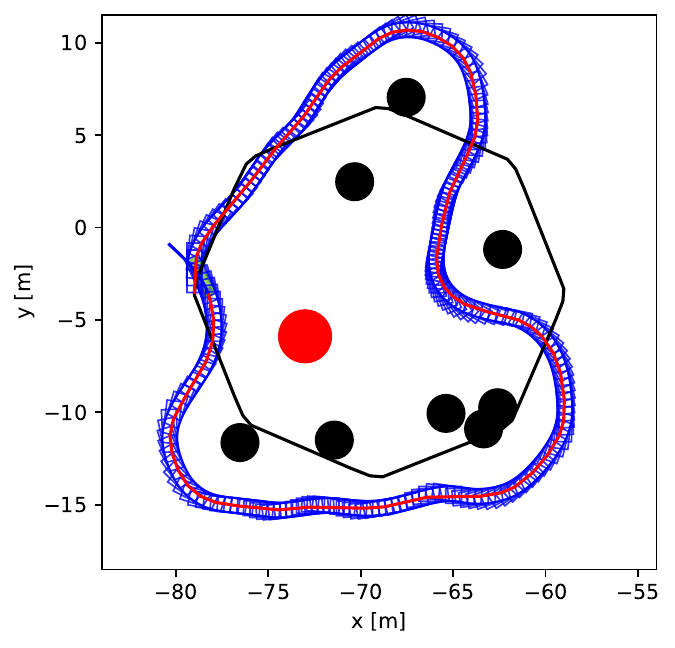}
          \vspace{-0.25in}
    \caption{MAC with PCA.}
  \end{subfigure}
  \caption{Trajectories of different schemes. The sensor is marked as a red circle and obstacles are marked as black circles. The global route is marked as a black line. Actual robot path is represented by a red line. The robot is marked as a blue box. }
  \label{traj2}
  \vspace{-0.2in}
\end{figure}

First, we consider the case of $K=1$ and $M=8$ to show how MCCA improves communication via control.
The communication efficiency and navigation time of different schemes are shown in Fig.~\ref{perf_los}. 
Each point in the figure is obtained by averaging $20$ random runs with $a_k\sim U(-75,-65)$ and $b_k\sim U(-8,0)$.
The results consistently demonstrate that, regardless of the number of obstacles, the proposed MCCA framework achieves the highest communication efficiency, with up to $14\%$ improvement compared to the other benchmark schemes.
Furthermore, the navigation time of MCCA is also the shortest as shown in Fig.~\ref{perf_los}b. 
This demonstrates that collision avoidance and data collection may not always be contradictory, in fact, they can be mutually beneficial. To gain insights into this result, the trajectories of all schemes in two particular examples are shown in Fig.~\ref{traj2}.
It can be seen that MCCA successfully avoids all obstacles and selects an inner path that simultaneously reduces the navigation time and increase the communication rate.
In contrast, the SCA scheme selects the outer path, which involves smoother movements, but positions the robot further from the sensor, resulting in a smaller communication throughput. 
Lastly, the MAC-PCA scheme leads to excessive navigation time due to its lack of consideration for robot shape.

\begin{figure*}[!t]
  \centering
  \begin{subfigure}[t]{0.255\textwidth}
    \centering
    \includegraphics[width=1\textwidth]{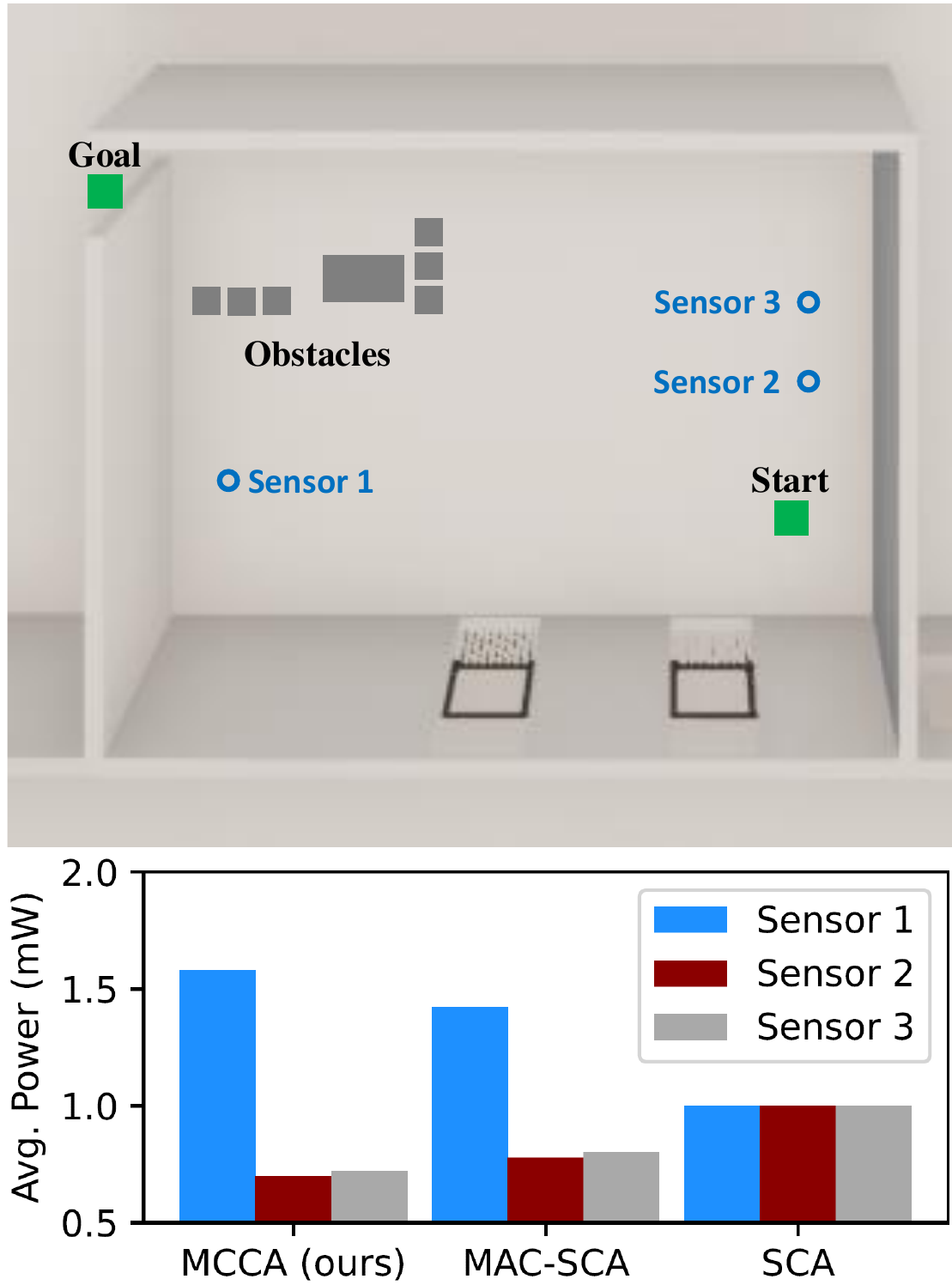}
    \caption{Scenario 1 and power profiles.}
  \end{subfigure}
  \begin{subfigure}[t]{0.24\textwidth}
      \centering
      \includegraphics[width=1\textwidth]{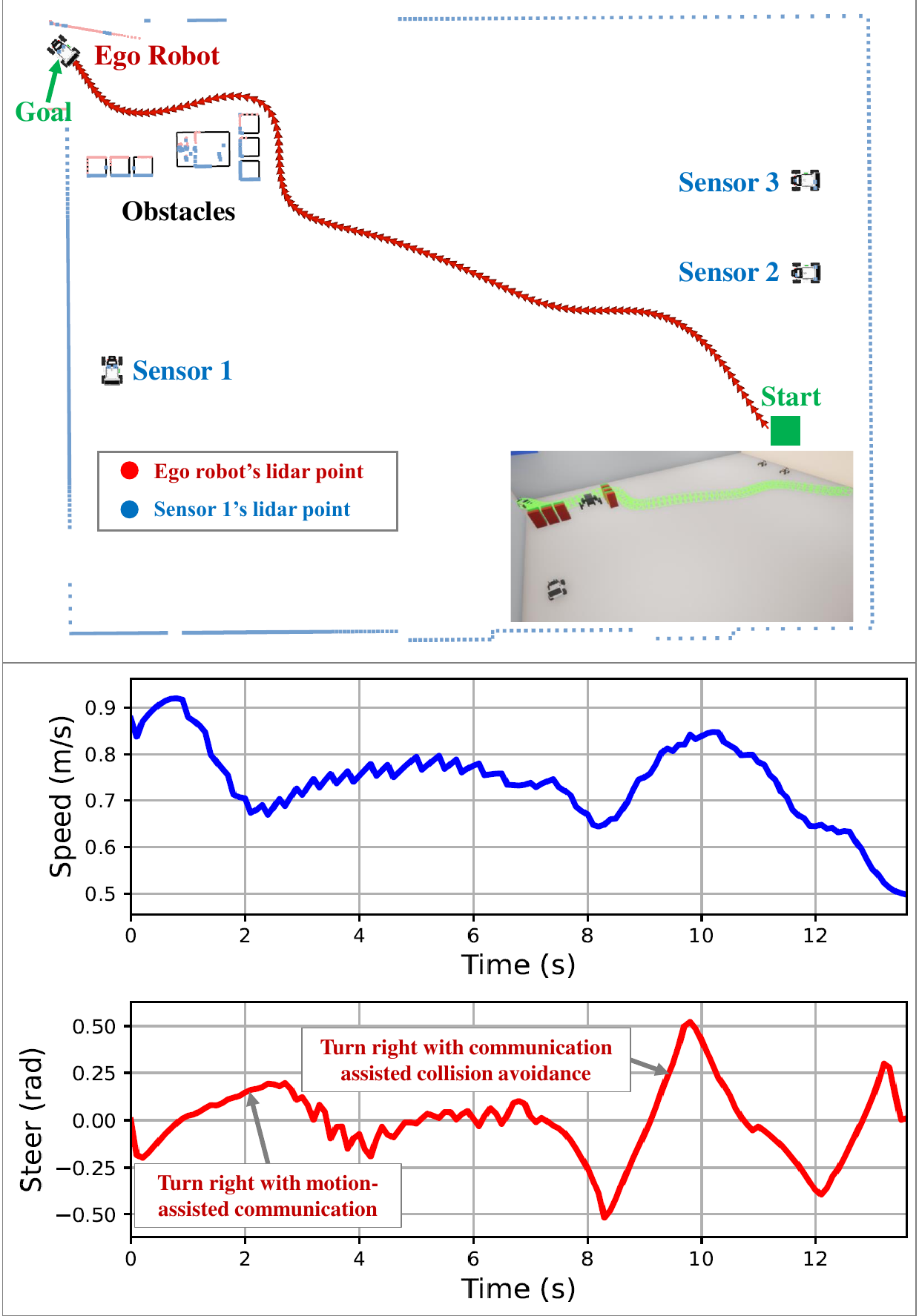}
      \caption{MCCA (ours).}
  \end{subfigure}
    \begin{subfigure}[t]{0.24\textwidth}
      \centering
      \includegraphics[width=1\textwidth]{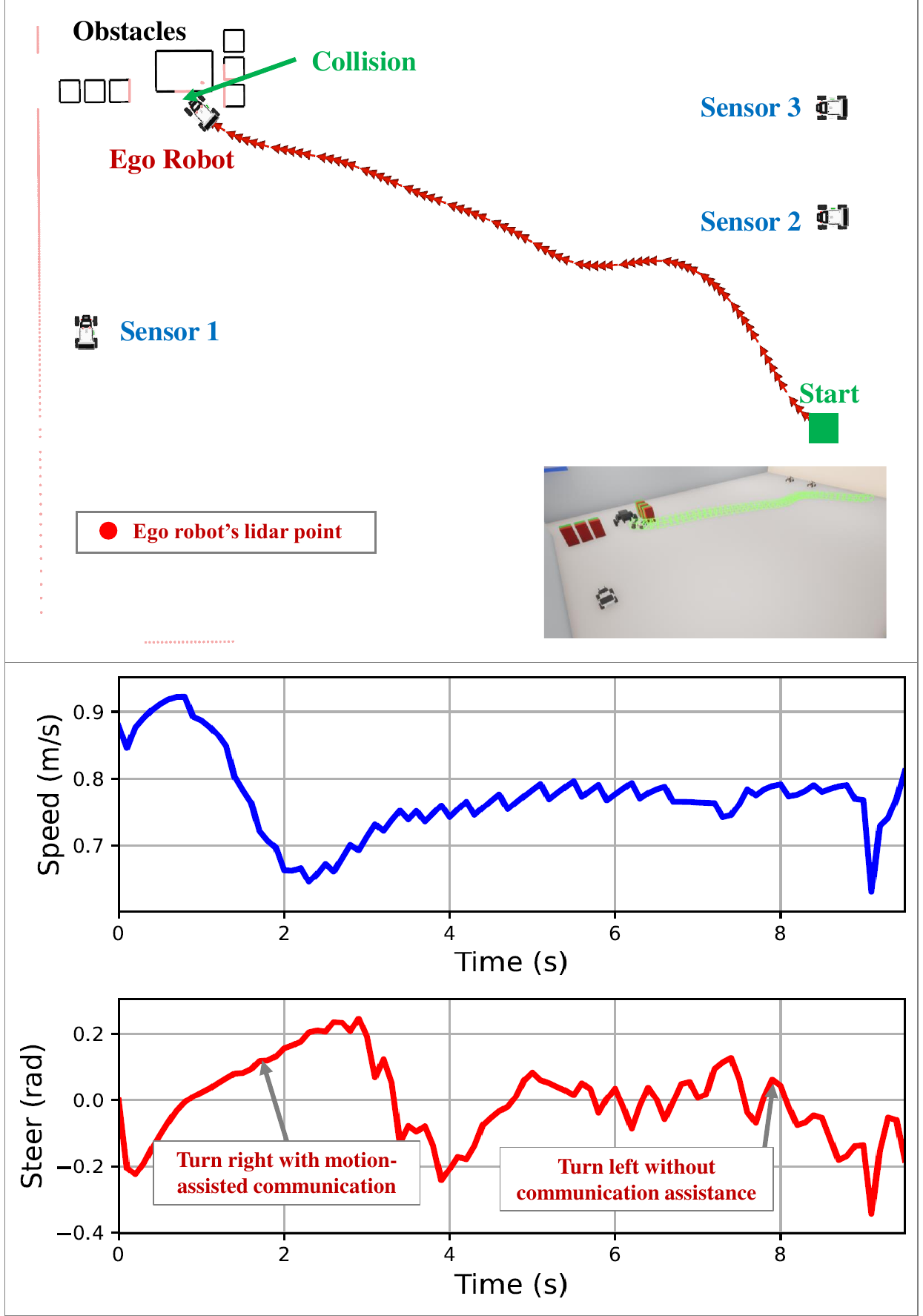}
      \caption{MAC with SCA.}
  \end{subfigure}
    \begin{subfigure}[t]{0.24\textwidth}
      \centering
      \includegraphics[width=1\textwidth]{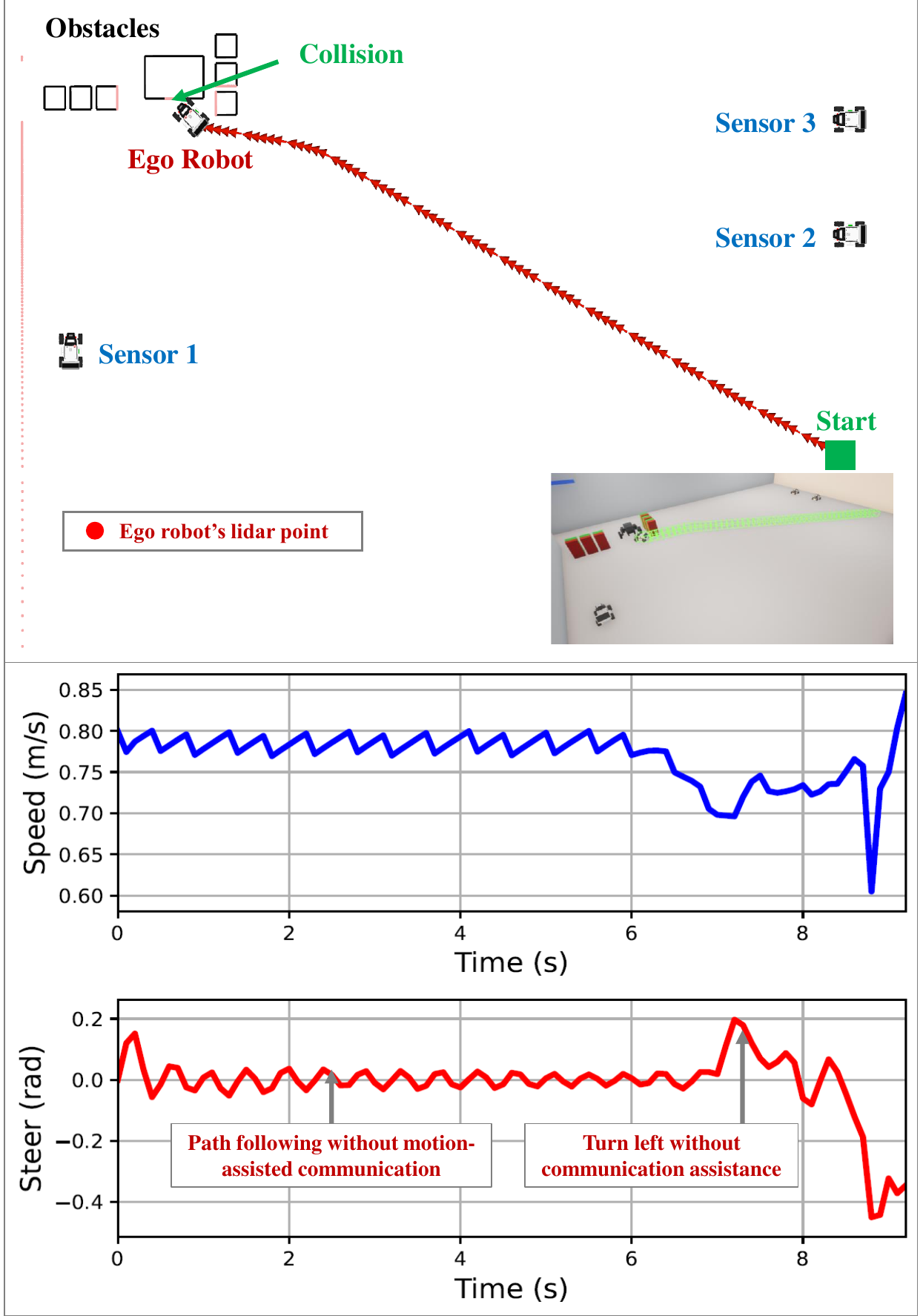}
      \caption{SCA.}
  \end{subfigure}
      \vspace{-0.1in}
  \caption{Simulated scenario 1 in CARLA and the sensor powers, robot trajectories, control commands of different schemes.}
    \vspace{-0.2in}
  \label{real_time}
\end{figure*}

\begin{figure}[!t]
  \centering
  \begin{subfigure}[t]{0.2\textwidth}
      \includegraphics[width=1\textwidth]{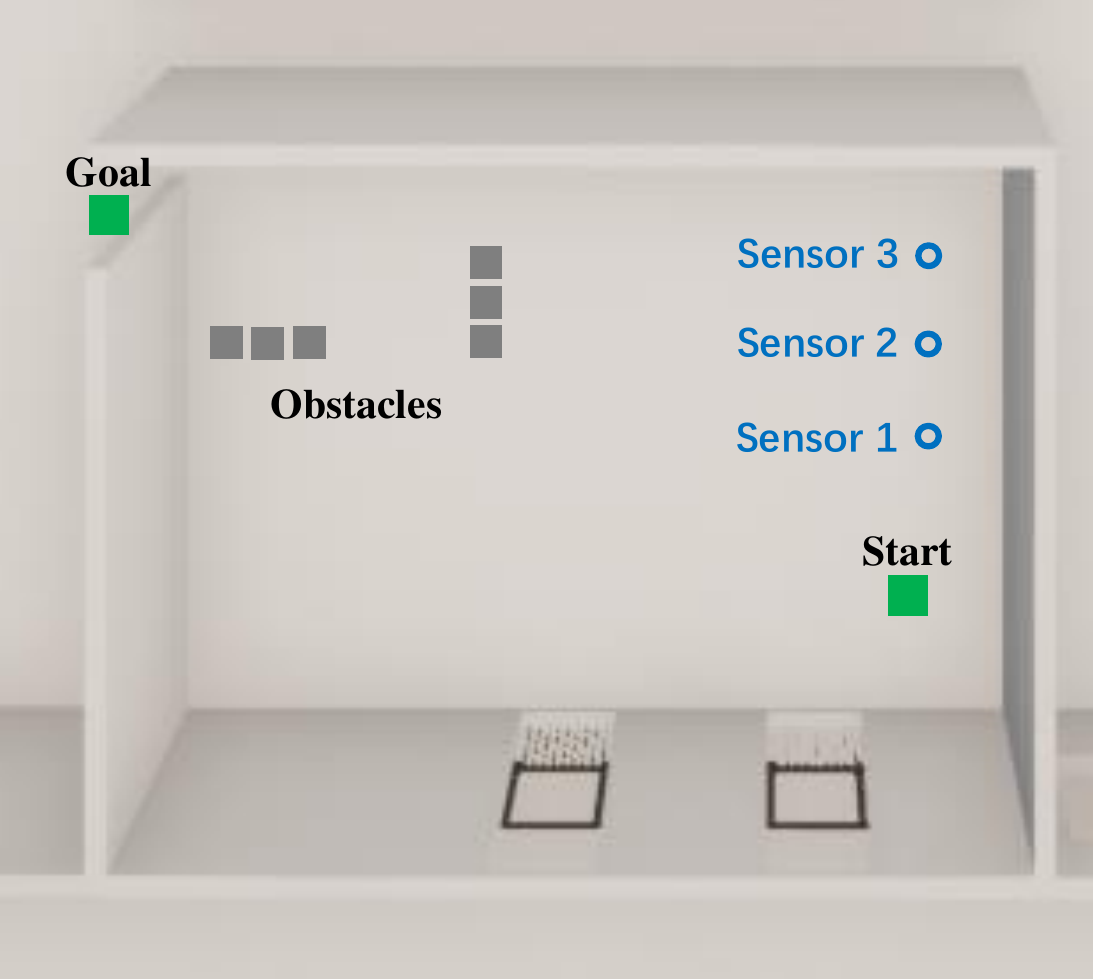}
      \caption{Scenario 2.}
  \end{subfigure}
  \begin{subfigure}[t]{0.25\textwidth}
    \includegraphics[width=1\textwidth]{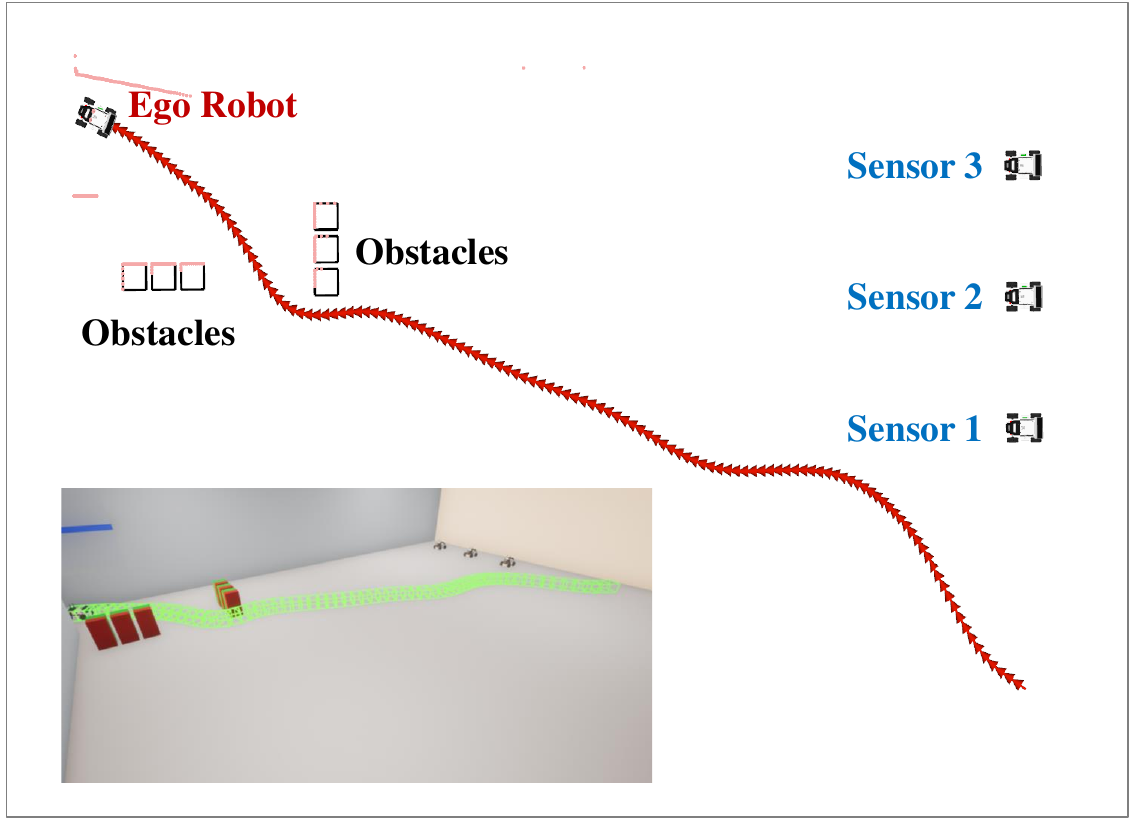}
    \caption{Robot trajectory.}
  \end{subfigure}
        \vspace{-0.1in}
  \caption{Simulated scenario 2 in CARLA.}
    \vspace{-0.2in}
\end{figure}

\begin{figure}[!t]
  \centering
    \begin{subfigure}[t]{0.24\textwidth}
    \includegraphics[width=1\textwidth]{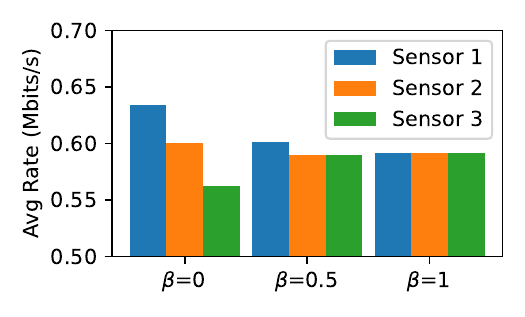}
    \vspace{-0.3in}
    \caption{Data rate.}
  \end{subfigure}
  \begin{subfigure}[t]{0.24\textwidth}
      \includegraphics[width=1\textwidth]{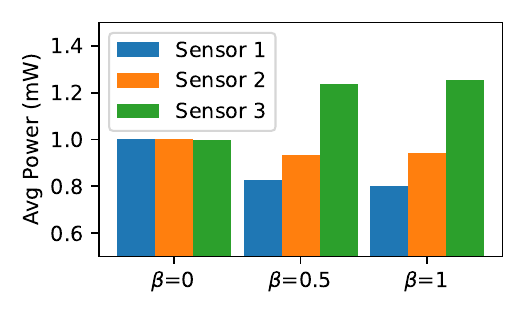}
      \vspace{-0.3in}
      \caption{Transmit power.}
  \end{subfigure}
        \vspace{-0.1in}
  \caption{Impact of $\beta$ on MCCA.}
          \vspace{-0.2in}
\end{figure}

Next, we consider the cases of $K=3$ and $M=7$ (comprising 6 cubes and 1 legged robot) in indoor navigation scenario 1, as shown in Fig. 4a, to demonstrate how MCCA adopts communication to improve control. 
To support realistic sensing and vehicle dynamics, we implement the proposed MCCA by ROS and verify its effectiveness in CARLA \cite{dosovitskiy2017carla}.
Specifically, external sensor 1 transmits its sensing data to the ego-robot to assist collision avoidance from $t=7$\,s to $t=10$\,s, with $\mathcal{A}=\{1\}$ and $D_0=0.7$\,Mbits/s.
The start (i.e., the corner of the room) and goal (i.e., the door of the room) positions are $(10,2)$ and $(2,6.3)$, respectively.
Sensors $1,2,3$ are located at $(2.5,3),(10,4),(10,5)$, respectively.
It can be observed from Figs. 4c and 4d that both MAC-SCA and SCA schemes turn left at about $t=8$\,s, leading to a collision with the legged robot.
This is because these schemes ignore the impact of communication on control, and the ego-robot fully depends on its own sensing (i.e., the red lidar points), which fails in detecting the legged robot behind the cubes due to occlusion. 
In contrast, under the proposed MCCA, the ego-robot slows down at $t=8$\,s, turns right at about $t=9$\,s, and successfully reaches the goal. 
This is because MCCA incorporates the communication-assisted collision avoidance into constraint \eqref{collision}. 
Hence, the blue points detected by sensor 1 complement the red points detected by the ego-robot, as shown in Fig. 4b, enabling the ego-robot to react to the legged robot in advance.
This corroborates Fig. 4a, where MCCA allocates the highest power to sensor 1, so as to satisfy the data-rate requirement of $D_0=0.7$\,Mbits/s for exchanging the sensing data. 
This also demonstrates the necessity of cross-layer optimization. 
Note that in Fig. 4, both MCCA and MAC-SCA schemes turn right to approach sensors 2 and 3 for motion-assisted communication. 

Lastly, we consider the case of $K=3$ and $M=6$ in indoor navigation scenario 2 in Fig. 5. Sensors $1,2,3$ are located at $(10,4),(10,5),(10,6)$, respectively, with $\mathcal{A}=\emptyset$.
The data rates and transmit powers of all sensors are shown in Fig. 6. 
It can be seen that the data-rates at all sensors are equal at $\beta=1$, which guarantees user fairness. 
This is realized by allocating the largest transmit power to the farthest sensor (i.e., sensor 3).
In contrast, at $\beta=0$, more data rates are allocated to the nearby sensors.
Finally, by adjusting the parameter $\beta$, the sum-rate and user fairness are balanced, which demonstrates the effectiveness of cross-layer optimization.

\vspace{-0.1in}
\section{Conclusion}\label{section5}

This paper presented the MCCA framework based on a newly derived MPC$^2$ formulation.
The MCCA adopted intertwined optimization of motion-assisted communication and communication-assisted motion.
A fast cross-layer optimization algorithm was proposed to jointly optimize the sensor powers and robot actions.
Experiments showed that MCCA achieves high-quality data gathering without any collision, and outperforms existing MAC, SCA, and PCA schemes.

\bibliographystyle{IEEEtran}
\bibliography{wcl_mcca}

\begin{thebibliography}{10}
\providecommand{\url}[1]{#1}
\csname url@samestyle\endcsname
\providecommand{\newblock}{\relax}
\providecommand{\bibinfo}[2]{#2}
\providecommand{\BIBentrySTDinterwordspacing}{\spaceskip=0pt\relax}
\providecommand{\BIBentryALTinterwordstretchfactor}{4}
\providecommand{\BIBentryALTinterwordspacing}{\spaceskip=\fontdimen2\font plus
\BIBentryALTinterwordstretchfactor\fontdimen3\font minus \fontdimen4\font\relax}
\providecommand{\BIBforeignlanguage}[2]{{%
\expandafter\ifx\csname l@#1\endcsname\relax
\typeout{** WARNING: IEEEtran.bst: No hyphenation pattern has been}%
\typeout{** loaded for the language `#1'. Using the pattern for}%
\typeout{** the default language instead.}%
\else
\language=\csname l@#1\endcsname
\fi
#2}}
\providecommand{\BIBdecl}{\relax}
\BIBdecl

\bibitem{zhou2019edge}
Z.~Zhou, X.~Chen, E.~Li, L.~Zeng, K.~Luo, and J.~Zhang, ``Edge intelligence: Paving the last mile of artificial intelligence with edge computing,'' \emph{Proc. IEEE}, vol. 107, no.~8, pp. 1738--1762, Aug. 2019.

\bibitem{ma2012tour}
M.~Ma, Y.~Yang, and M.~Zhao, ``Tour planning for mobile data-gathering mechanisms in wireless sensor networks,'' \emph{IEEE Trans. Veh. Technol.}, vol.~62, no.~4, pp. 1472--1483, May 2012.

\bibitem{chen2021ugv}
E.~Chen, P.~Wu, Y.-C. Wu, and M.~Xia, ``{UGV}-assisted wireless powered backscatter communications for large-scale {IoT} networks,'' \emph{IEEE Trans. Wireless Commun.}, vol.~21, no.~5, pp. 3147--3161, May 2021.

\bibitem{licea2024when}
D.~B. Licea, M.~Ghogho, and M.~Saska, ``When robotics meets wireless communications: An introductory tutorial,'' \emph{Proc. IEEE}, vol. 112, no.~2, pp. 140--177, Feb. 2024.

\bibitem{wang2019backscatter}
S.~Wang, M.~Xia, and Y.-C. Wu, ``Backscatter data collection with unmanned ground vehicle: Mobility management and power allocation,'' \emph{IEEE Trans. Wireless Commun.}, vol.~18, no.~4, pp. 2314--2328, Apr. 2019.

\bibitem{zhou2020learning}
L.~Zhou, Y.~Hong, S.~Wang, R.~Han, D.~Li, R.~Wang, and Q.~Hao, ``Learning centric wireless resource allocation for edge computing: Algorithm and experiment,'' \emph{IEEE Trans. Veh. Technol.}, vol.~70, no.~1, pp. 1035--1040, Jan. 2020.

\bibitem{guo2021uav}
Y.~Guo, C.~You, C.~Yin, and R.~Zhang, ``{UAV} trajectory and communication co-design: Flexible path discretization and path compression,'' \emph{IEEE J. Sel. Areas Commun.}, vol.~39, no.~11, pp. 3506--3523, Nov. 2021.

\bibitem{ali2018motion}
U.~Ali, H.~Cai, Y.~Mostofi, and Y.~Wardi, ``Motion-communication co-optimization with cooperative load transfer in mobile robotics: An optimal control perspective,'' \emph{IEEE Trans. Control Netw. Syst.}, vol.~6, no.~2, pp. 621--632, Jun. 2018.

\bibitem{yan2023communication}
J.~Yan, L.~Zhang, X.~Yang, C.~Chen, and X.~Guan, ``Communication-aware motion planning of {AUV} in obstacle-dense environment: A binocular vision-based deep learning method,'' \emph{IEEE Trans. Intell. Transp. Syst.}, vol.~24, no.~12, pp. 14\,927--14\,943, Dec. 2023.

\bibitem{jasontits}
S.~Zhang, S.~Wang, S.~Yu, J.~Yu, and M.~Wen, ``Collision avoidance predictive motion planning based on integrated perception and {V2V} communication,'' \emph{IEEE Trans. Intell. Transp. Syst.}, vol.~23, no.~7, pp. 9640--9653, Jul. 2022.

\bibitem{zhang2020optimization}
X.~Zhang, A.~Liniger, and F.~Borrelli, ``Optimization-based collision avoidance,'' \emph{IEEE Trans. Control Syst. Technol.}, vol.~29, no.~3, pp. 972--983, May 2021.

\bibitem{han2023rda}
R.~Han, S.~Wang, S.~Wang, Z.~Zhang, Q.~Zhang, Y.~C. Eldar, Q.~Hao, and J.~Pan, ``{RDA}: An accelerated collision free motion planner for autonomous navigation in cluttered environments,'' \emph{IEEE Robot. Autom. Lett.}, vol.~8, no.~3, pp. 1715--1722, Mar. 2023.

\bibitem{evangelista2019fairness}
J.~V. Evangelista, Z.~Sattar, G.~Kaddoum, and A.~Chaaban, ``Fairness and sum-rate maximization via joint subcarrier and power allocation in uplink {SCMA} transmission,'' \emph{IEEE Trans. Wireless Commun.}, vol.~18, no.~12, pp. 5855--5867, Dec. 2019.

\bibitem{flcav}
S.~Wang, C.~Li, D.~W.~K. Ng, Y.~C. Eldar, H.~V. Poor, Q.~Hao, and C.~Xu, ``Federated deep learning meets autonomous vehicle perception: Design and verification,'' \emph{IEEE Netw.}, vol.~37, no.~3, pp. 16--25, May/Jun. 2023.

\bibitem{diamond2016cvxpy}
S.~Diamond and S.~Boyd, ``{CVXPY}: A python-embedded modeling language for convex optimization,'' \emph{J. Mach. Learn. Res.}, vol.~17, no.~1, pp. 2909--2913, 2016.

\bibitem{sun2016majorization}
Y.~Sun, P.~Babu, and D.~P. Palomar, ``Majorization-minimization algorithms in signal processing, communications, and machine learning,'' \emph{IEEE Trans. Signal Process.}, vol.~65, no.~3, pp. 794--816, Feb. 2017.

\bibitem{dosovitskiy2017carla}
A.~Dosovitskiy, G.~Ros, F.~Codevilla, A.~Lopez, and V.~Koltun, ``Carla: An open urban driving simulator,'' in \emph{Conf. Robot Learn. (CoRL)}.\hskip 1em plus 0.5em minus 0.4em\relax PMLR, 2017, pp. 1--16.

\end{thebibliography}

\end{document}